\documentclass{bmvc2k}
\usepackage{amsfonts}
\usepackage{multirow}
\usepackage{amssymb}
\usepackage{algorithm}
\usepackage{verbatim}
\usepackage[noend]{algpseudocode}
\usepackage{footmisc}
\renewcommand{\thefootnote}{\fnsymbol{footnote}}

\newcommand\blfootnote[1]{%
  \begingroup
  \renewcommand\thefootnote{}\footnote{#1}%
  \addtocounter{footnote}{-1}%
  \endgroup
}

\title{SuperStyleNet: Deep Image Synthesis with Superpixel Based Style Encoder}

\addauthor{Jonghyun Kim}{jhkim.ben@gmail.com}{1}
\addauthor{Gen Li}{li.gen@ed.ac.uk}{2}
\addauthor{Cheolkon Jung$^\dagger$}{zhengzk@xidian.edu.cn}{3}
\addauthor{Joongkyu Kim$^\dagger$}{jkkim@skku.edu}{1}

\addinstitution{ \small
 Department of Electrical and Computer Engineering\\
 Sungkyunkwan University\\
 Suwon, South Korea
}
\addinstitution{ \small
 School of Informatics,\\
 University of Edinburgh,\\
 Edinburgh, UK
}
\addinstitution{ \small
 School of Electronic Engineering,\\
 Xidian University,\\
  Xian, Shaanxi, China
}

\runninghead{KIM, \etal.}{SuperStyleNet for Semantic Image Synthesis}

\def\etal{\emph{et al}\bmvaOneDot}

\newif\ifdrafting
\draftingtrue 

\ifdrafting
    \newcommand{\LG} [1] {\textcolor{red}{[LG: #1]}}
    \newcommand{\JH} [1] {\textcolor{blue}{[JH: #1]}}
\else
    \newcommand{\LG} [1] {}
    \newcommand{\JH} [1] {}
\fi

\begin{document}

\maketitle
\vspace{-0.6cm}
\begin{abstract}
   Existing methods for image synthesis utilized a style encoder based on stacks of convolutions and pooling layers to generate style codes from input images. However, the encoded vectors do not necessarily contain local information of the corresponding images since small-scale objects are tended to "wash away" through such downscaling procedures. In this paper, we propose deep image synthesis with superpixel based style encoder, named as SuperStyleNet. First, we directly extract the style codes from the original image based on superpixels to consider local objects. Second, we recover spatial relationships in vectorized style codes based on graphical analysis. Thus, the proposed network achieves high-quality image synthesis by mapping the style codes into semantic labels. Experimental results show that the proposed method outperforms state-of-the-art ones in terms of visual quality and quantitative measurements. Furthermore, we achieve elaborate spatial style editing by adjusting style codes. The codes are available at: \url{https://github.com/BenjaminJonghyun/SuperStyleNet}\blfootnote{$\dagger$ Corresponding authors}
\end{abstract}

\begin{figure}[!h]
    \centering
    \includegraphics[width=\linewidth]{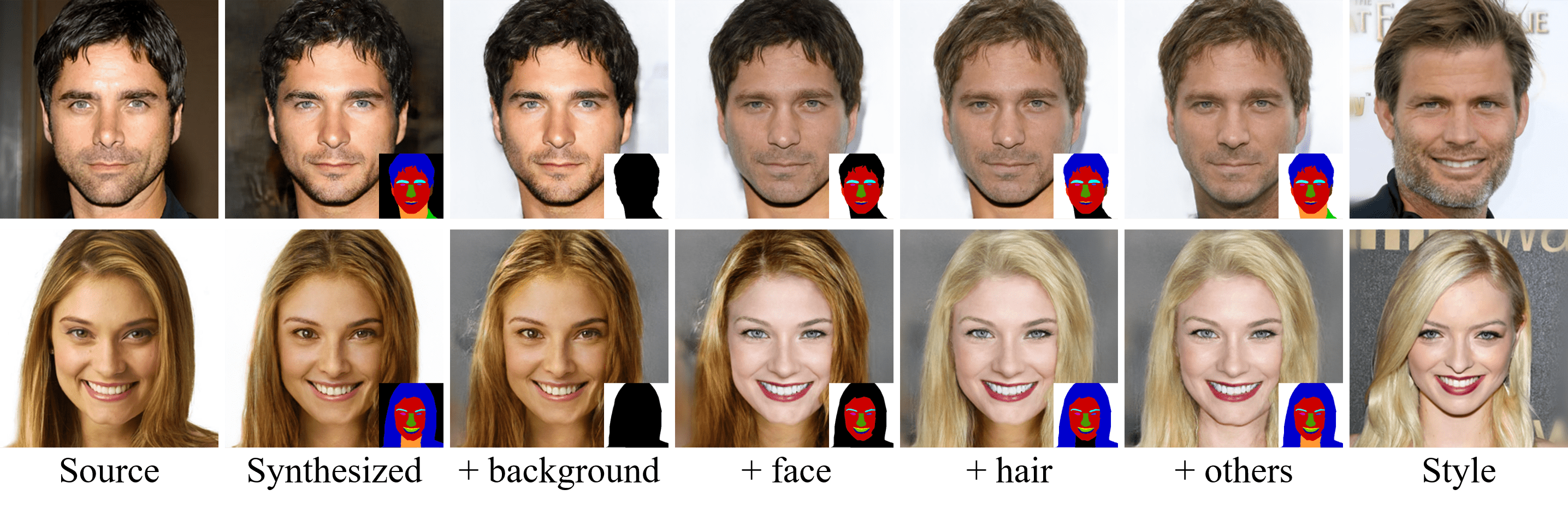}
    \vspace{-0.8cm}
    \caption{Face image synthesis by the proposed network. The synthesized images are generated with segmentation masks of source images and their style information. In other manipulated images, the style is controlled by replacing style codes in given segmentation masks from source to style images.}
    \vspace{-0.5cm}
    \label{Main}
\end{figure}

\vspace{-0.8cm}
\section{Introduction} 
\vspace{-0.3cm}
\label{sec:intro}
The goal of conditional image synthesis is to generate photo-realistic images conditioning on certain input data. Recent methods utilized neural networks to generate realistic images from other images, latent codes, edge maps, or pose key points \cite{brock2018large, isola2017image, wang2018high, zhang2017stackgan, hwang2020fairfacegan, tang2020bipartite, na2019miso, alharbi2019latent}. Especially, our interest lies in using semantic layouts to assist conditional image synthesis \cite{chen2017photographic, isola2017image, qi2018semi, wang2018high}, which generates photo-realistic images from semantic masks. In addition, it can manipulate semantic information in images, such as context generation and image editing, by controlling the segmentation masks. Although these methods generate high-quality images, it cannot freely control image styles since the segmentation masks are only fed to their networks. To achieve it, SPADE \cite{park2019semantic} and SelectionGAN \cite{tang2019multi} utilized a style image to extract style information by using an encoder network. Then, a decoder network conducts style reconstruction in semantic layouts by referring to the style information, and generates a synthesized image similar to the style image. However, the style information is characterized to represent large-scale objects since downscaling by convolutions and pooling layers tends to gradually "wash away" small-scale objects in feature maps. It indicates that the generator shows a bias toward large-scale objects rather than smaller ones during the learning process. Thus, it is difficult to conduct the style reconstruction of the local information. To 
equally contain local and global information, current methods \cite{tang2020local, zhu2020sean} modified the existing encoder or decoder to reconstruct local information from downscaled feature maps or encoded style vectors. However, undesirable effects appear in the results by reconstructing local information from high-level features (i.e., downscaled feature maps and encoded style vectors). In this case, there exists little style information of the small-scale objects in the high-level features. Thereby, it is difficult to expect that the local information can be well reconstructed when the decoder conducts style mapping in the local ones.

To tackle this issue, a straightforward strategy is to encode the style information from the original image while maintaining the original scale. However, it brings a large amount of parameters in the style encoder. Therefore, we propose Superpixel based Parameter-free Style Encoding (SPSE) to directly encode the original image into the style vector. We apply SLIC superpixel segmentation \cite{achanta2012slic} to each segmentation mask for generating the style vector per semantic region. It can provide a different view from the existing style-encoding methods \cite{richardson2020encoding, karras2019style, liu2019few}. Furthermore, we analyze hidden representation of each node between nearest pixels to inject their spatial relationships to the style vector with a Graphical Self-Attention Strategy (GSAS) because compressing an image to the style vector based on superpixel wipes the spatial relationships off. Consequently, the proposed network is capable of synthesizing a high-quality image by considering both local and global information.

We provide extensive experiments to prove the effectiveness of the proposed method on challenging datasets: CelebAMask-HQ \cite{lee2020maskgan}, Cityscapes \cite{cordts2016cityscapes} and CMP Facades \cite{Tylecek13facade}. We evaluate the performance of the proposed network in terms of various metrics. Compared with existing methods, our contributions can be summarized as follows:
\begin{itemize}
    \item We propose superpixel based parameter-free style encoding, called as SPSE, to regularly encode local and global style information of the input image into the style vector with a parameter-free operation.
    \item We provide GSAS to compensate for loss of the spatial relationships between neighbor pixels of the encoded style vector. This strategy uses graph based the self-attention method to define them.
    \item The proposed network achieves better performance in image synthesis than state-of-the-art ones in terms of visual quality and quantitative measurements.
\end{itemize}

\vspace{-0.6cm}
\section{Related Work}
\label{sec:related}
\vspace{-0.3cm}
\textbf{Image-to-Image Translation} is to learn a parametric mapping from input to output. Isola \textit{et al.} \cite{isola2017image} first proposed image-to-image translation with a conditional generative adversarial network (GAN) to translate the source to target domains. This task has been widely extended by many researchers. Huang \textit{et al.} \cite{huang2018multimodal} and Alharbi \textit{et al.} \cite{alharbi2019latent} proposed multi-modal image-to-image translation in an unsupervised way. Moreover, various methods \cite{choi2018stargan, almahairi2018augmented, choi2020stargan} were proposed to conduct multi-domain translation. As a specific task of the image-to-image translation, semantic image synthesis \cite{bau2020semantic, park2019semantic, gu2019mask, lee2020maskgan, zhu2020sean, sushko2020you} has become much popular in terms of unconstrained image control. To be specific, SPADE \cite{park2019semantic} proposed spatially-adaptive denormalization to preserve semantic information of output images similar to input semantic layouts. Furthermore, SEAN \cite{zhu2020sean} improved SPADE by regionally normalizing parameters to control the style of each semantic region individually. Although these methods showed outstanding performance on semantic image synthesis, there exists shortcomings in the process of style information encoding. The repeated downscalings with convolution layers in the style encoding erase features of local information. To overcome this shortcoming, the proposed method equally encodes local and global information into the style vector based on superpixels.
\\ \quad
\\ \textbf{Style Encoding} is essential in image synthesis to extract style information from reference images. Existing methods \cite{gatys2015neural, johnson2016perceptual} utilized a VGG \cite{simonyan2014very} network pre-trained on image classification \cite{deng2009imagenet} to obtain style features. Unlike these methods, AdaIN \cite{huang2017arbitrary} inferred affine parameters from the style image to map input parameters into the style space, which enabled arbitrary style transfer in real-time. On account of its easy application and outstanding performance, this method was adopted in various tasks \cite{huang2018multimodal, kotovenko2019content, karras2019style}. Furthermore, Park \textit{et al.} \cite{park2019semantic} and Zhu \textit{et al.} \cite{zhu2020sean} improved AdaIN to achieve semantic manipulation by normalizing the affine parameters in spatial or semantic regions, but neural networks are still indispensable for the style encoding. Compared with the aforementioned methods, the proposed method provides a new perspective in terms of non-parametric style encoding.

\vspace{-0.3cm}
\section{Proposed Method}
\vspace{-0.3cm}
We aim to regularly extract local and global style information per semantic mask from the reference image. The encoded style is scattered into its corresponding masks to synthesize photo-realistic images. To implement this concept, we first introduce the superpixel based parameter-free style encoding (SPSE) and the graphical self-attention strategy (GSAS). Then, we describe the overall network architecture of a superpixel based style encoding network, named as SuperStyleNet.

\begin{figure}
    \centering
    \includegraphics[width=\linewidth]{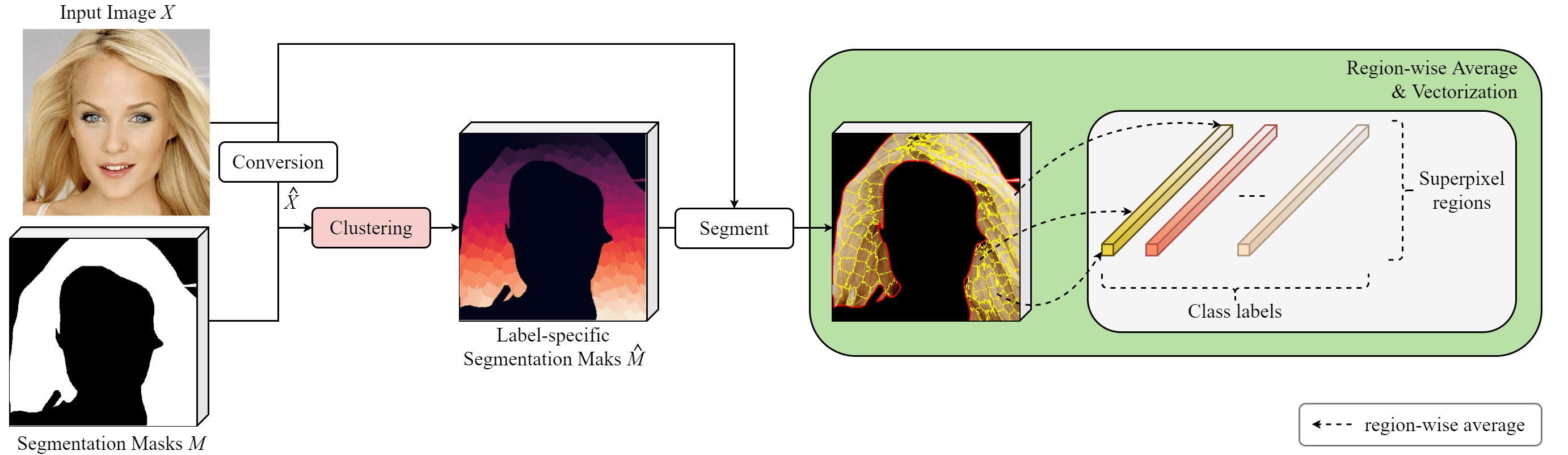}
    \vspace{-0.3cm}
    \caption{Illustration of the proposed Superpixel based Parameter-free Style Encoding (SPSE). To extract style codes of a specific semantic mask, we convert the input image into the five-dimensional space and cluster it in the semantic mask into superpixels. Thereafter, pixel values in each superpixel are averaged to obtain a style code.}
    \vspace{-0.3cm}
    \label{SPSE}
\end{figure}

\subsection{Superpixel based Parameter-free Style Encoding}
A superpixel can be defined as a group of pixels that share similar visual characteristics. SLIC \cite{achanta2012slic} is a representative superpixel method, which clusters pixels based on color similarity and proximity. To be specific, the clustering procedure is conducted in a five-dimensional space $[l\;a\;b\;x\;y]$, where $[l\;a\;b]$ is the pixel color vector in CIELAB color space and $[x\;y]$ is the pixel position. During the clustering, each pixel in the input image is assigned to $k$ superpixels by calculating the $l2$ distance between each initial cluster center and its neighborhood. Inspired by this method, we propose superpixel based parameter-free style encoding (SPSE) as illustrated in Fig. \ref{SPSE}, which allocates an input image to $k$ desired superpixels, and converts them to the style code in the color space. 

Let $M \in \mathbb{I}^{H \times W \times L}$ be a semantic segmentation mask given the number of class labels $L$, where $\mathbb{I}$ is a set of integers consisting of either "$1$" or "$0$", and $H \times W$ is the image size. Before clustering the RGB input image $X\in \mathbb{R}^{H\times W\times 3}$, we convert it into the five-dimensional space $\hat{X}\in \mathbb{R}^{H\times W\times 5}$, and superpixel centers are initialized with an uniform distribution \cite{irving2016maskslic}. After that, we repeat the clustering in each segmentation mask by SLIC \cite{achanta2012slic} to obtain superpixels per semantic label:
\begin{equation}
    SP = \textrm{Clustering}(\hat{X}[M==1]),
\end{equation}
where $SP \in \mathbb{R}^{L\times K}$ is a set of superpixels in each class label with the number of superpixel centers $K$. Then, each superpixel region is converted as label-specific segmentation masks $\hat{M} \in \mathbb{I}^{H\times W\times L\times K}$. To obtain the style information from the original image $X$, we segment pixels from the input image referring to $\hat{M}$, and take average on pixel values per label-specific semantic region as follows:
\begin{equation}
    SC= \mathbb{E} [X[\hat{M}==1]],
\end{equation}
where $SC\in\mathbb{R}^{L\times K\times 3}$ is style codes. Then, it is reshaped to $L\times 3K$, and interpolated with the desired length $N$ of the style code in axis "$1$".

\vspace{-0.3cm}
\subsection{Graphical Self-Attention Strategy}
Graph-based self-attention was proposed in \cite{velivckovic2017graph}, which computes the hidden representation of each node in graphs and is used in natural language processing (NLP) tasks. Due to its effective nodal analysis and attention mechanism, this method was adapted to interpret spatial representations, i.e., point cloud semantic segmentation \cite{wang2019graph}, medical image segmentation \cite{oktay2018attention}, human pose estimation \cite{zhao2019semantic}, and trajectory forecasting \cite{kosaraju2019social}.

\begin{figure}
    \centering
    \includegraphics[width=\linewidth]{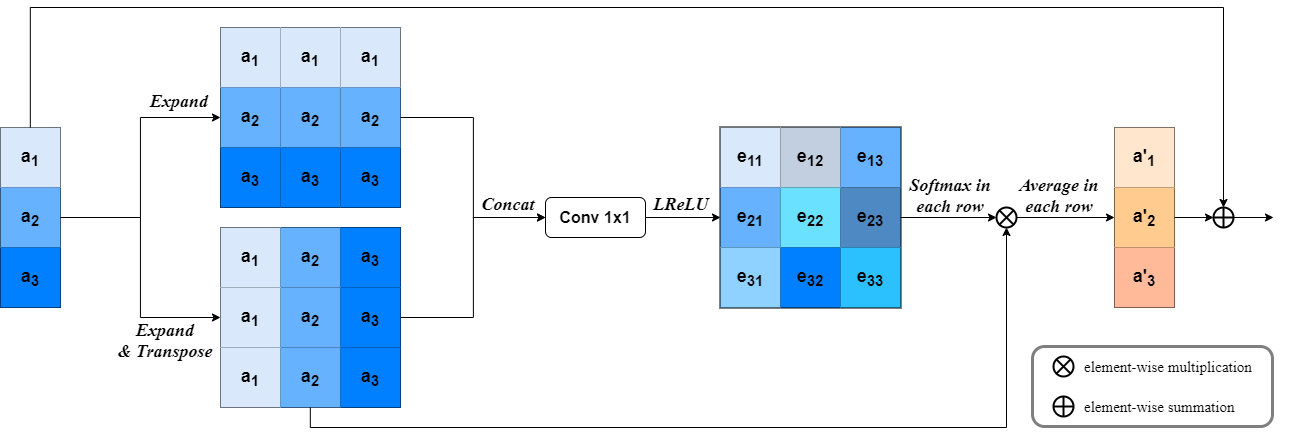}
    \vspace{-0.3cm}
    \caption{Flow of the proposed graphical self-attention strategy. For better understanding, we exemplify the style vector is set to be the length 3 given a specific class label. "Concat" and "LReLU" are notated as the concatenation and LeakyReLU operation, respectively.}
    \label{GSAS}
\end{figure}

Similarly, we propose a graphical self-attention strategy (GSAS), as described in Fig. \ref{GSAS}, to inject spatial hidden representations to the style vector. Let $\textbf{a}_{i} = \{a_1,a_2,...,a_N\}$ be the style vector given a specific class label, where $N$ is the length of the style vector. In order to convert the style vector to the spatial domain, the style vector \textbf{a} is expanded as a $N\times N$ matrix $\textbf{a}_{ij}$, where $i,j \in \{1, 2, ..., N\}$. After that, $\textbf{a}_{ij}$ is concatenated with its transposed matrix $\textbf{a}_{ij}^{T}$, then a $1\times 1$ convolution is used to analyze relationships between style components in the style vector and reduce the channel size to 1. Thereafter, we apply the LeakyReLU nonlinearity with $\alpha = 0.2$. Thus, correlation coefficients can be described as:
\begin{equation}
    \textbf{e}_{ij} = \textrm{LeakyReLU}(\textbf{W}([\textbf{a}_{ij}\parallel \textbf{a}_{ij}^{T}])),
    \label{coefficient}
\end{equation}
where \textbf{W} is a $1\times 1$ convolution, and $\parallel$ represents concatenation operation. Following the Eq. \ref{coefficient}, we obtain a correlation coefficient between $i$ and $j$-th style components. To convert the coefficients to comparable scores across different style components, we normalize them across all $j$ using the softmax function as follows:
\begin{equation}
    \textbf{s}_{ij} = \frac{exp(\textbf{e}_{ij})}{\sum_{k\in N}exp(\textbf{e}_{ik})}.
\end{equation}
Then, we conduct pixel-wise multiplication between $\textbf{a}_{ij}^{T}$ and $\textbf{s}_{ij}$ and averaging across all $j$ to obtain a new style vector $\textbf{a}_{i}'$ considered all $i\neq j$-th style components. Before feeding the style vector to the generator, the element-wise summation is operated to aggregate both style vectors $\textbf{a}_{i}$ and $\textbf{a}_{i}'$.

\subsection{Network Structure and Learning Objective}
\begin{figure}
    \centering
    \includegraphics[width=\linewidth]{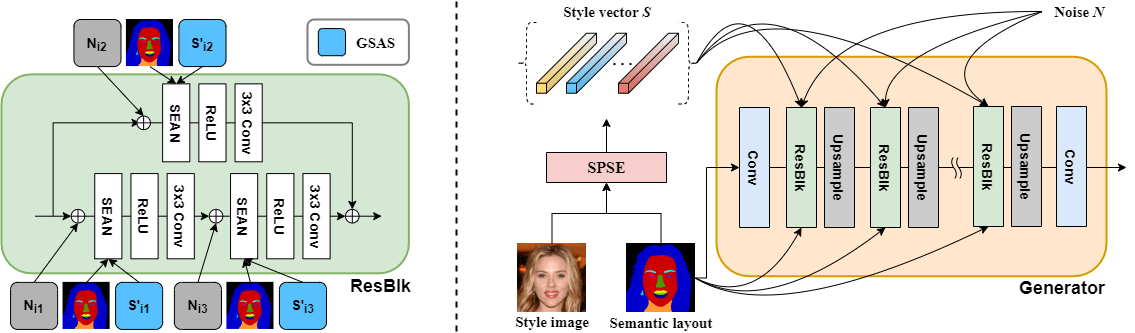}
    \vspace{-0.4cm}
    \caption{Illustration of the proposed SuperStyleNet. (\textit{left}) Structure of a residual block embedded in SuperStyleNet. (\textit{right}) Overview of SuperStyleNet, which contains a series of the residual blocks with $2\times$ upsampling (nearest) layers.}
    \vspace{-0.3cm}
    \label{structure}
\end{figure}

In this paper, we concentrate on semantic image synthesis and spatial style editing. To achieve them, we follow network structures in \cite{zhu2020sean, park2019semantic} but exclude the style encoder as illustrated in Fig. \ref{structure}, which employs several ResNet blocks \cite{he2016deep} with upsampling layers. In each ResNet block, the affine parameters are learned by embedding SEAN 
\cite{zhu2020sean} to scatter the style vector into the corresponding semantic layouts. Furthermore, we use the semantic layouts as the input of the generator, and feed the concatenation of the layouts and the synthesized images into the multi-scale discriminator \cite{wang2018high}.

With the help of SPSE, the embedding of the style encoder is not required since SPSE utilizes superpixels in the color domain to generate the style vector with non-parametric operation. Thus, the generator and discriminator excluding the style encoder are only considered in the training process. As the loss functions for the generator, we use perceptual loss \cite{johnson2016perceptual}, feature matching loss \cite{wang2018high}, and conditional adversarial loss \cite{zhu2020sean}. For the discriminator, the hinge loss term \cite{miyato2018spectral, zhang2019self} is adopted. We provide further details of the network structure and objectives in the supplemental materials. 

\section{Experiments}
\vspace{-0.2cm}
\subsection{Implementation Details}
\vspace{-0.2cm}
In our experiments, following the experimental conditions in \cite{zhu2020sean}, the style length $N$ is set to 512, and the class labels are decided by the number of class in datasets. We apply spectral normalization \cite{miyato2018spectral} to both the generator and discriminator, and a synchronized version of batch normalization is applied to SEAN layers of the residual block. Following a two time-scale update rule (TTUR) \cite{heusel2017gans}, the learning rates are set to $0.0001$ and $0.0004$ for the generator and discriminator respectively. Adam optimizer \cite{kingma2014adam} is adopted with $\beta_{1}=0$ and $\beta_{2}=0.999$.

We train and test SuperStyleNet on the segmentation datasets: CelebAMask-HQ \cite{lee2020maskgan}, Cityscapes \cite{cordts2016cityscapes} and CMP Facades \cite{Tylecek13facade}. 1) CelebAMask-HQ is a face image dataset containing 30,000 face images with 19 segmentation labels, which is split into 28,000 and 2000 images for train and test sets, respectively. 2) Cityscapes contains 3500 images annotated with 35 segmentation labels. For this dataset, the train and test set sizes are 3,000 and 500, respectively. 3) CMP Facades consists of 500 facade images with 12 segmentation labels. In this dataset, 400 and 100 images are utilized as train and test sets, respectively. All images are resized to $256\times256$ in both training and testing.

\begin{figure}[b]
    \centering
    \vspace{-0.3cm}
    \includegraphics[width=\linewidth]{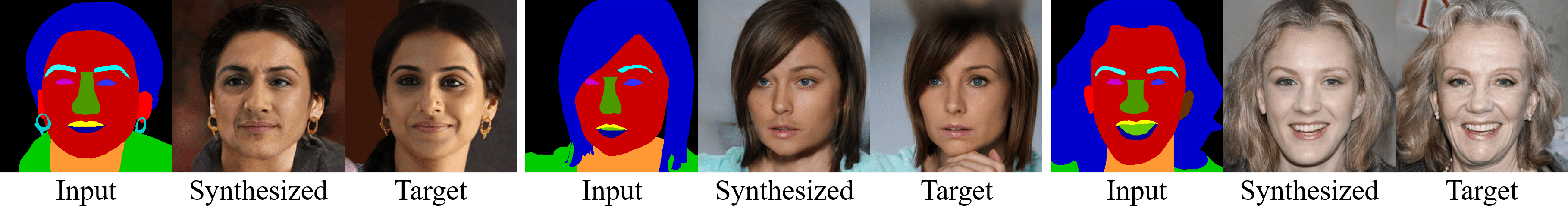}
    \vspace{-0.8cm}
    \caption{Failure cases of image synthesis. The first and second examples represent gender mismatch, while the last one is for age mismatch.}
    \vspace{-0.3cm}
    \label{failure}
\end{figure}
\vspace{-0.2cm}
\subsection{Evaluation Metrics}
\vspace{-0.2cm}
Following previous works \cite{wang2018high, park2019semantic, zhu2020sean}, we perform semantic segmentation on synthesized images to quantify how well the predicted segments match ground-truth. Specifically, BiSeNet \cite{yu2020bisenet} is applied to the synthesized images to infer semantic segmentation results, and pixel-wise accuracy (pix acc) and mean intersection-over-union (mIoU) are utilized as its evaluation metrics. Furthermore, we compare SuperStyleNet with these state-of-the-art methods by adopting peak signal-to-noise ratio (PSNR), normalized root mean square error (NRMSE), Fr{\'e}chet Inception Distance (FID) \cite{heusel2017gans}, and learned perceptual image patch similarity (LPIPS) \cite{zhang2018unreasonable}.

\begin{table}[t]
\centering
\caption{Quantitative comparison on semantic segmentation and generation performance. Higher mIoU, higher pixel acc, and lower FID indicate better performance.}
\resizebox{\textwidth}{!}{\begin{tabular}{l||c|c|c||c|c|c||c|c|c}
\hline
\multicolumn{1}{c||}{\multirow{2}{*}{Method}} & \multicolumn{3}{|c||}{CelebAMask-HQ}               & \multicolumn{3}{c||}{Cityscapes} & \multicolumn{3}{c}{CMP Facades}                  \\ \cline{2-10} 
\multicolumn{1}{c||}{}                        & mIoU           & pix acc        & FID            & mIoU     & pix acc    & FID     & mIoU           & pix acc        & FID             \\ \hline \hline
Ground Truth                                  & 70.05          & 36.75          & 8.71           & 50.46    & 91.50      & 34.20   & 36.75          & 71.28          & 75.07           \\ \hline \hline
Pix2PixHD                                     & 73.15          & 95.22          & 27.45          & 49.21    & 91.18      & 104.39   & 40.83          & 72.19          & 156.91          \\ \hline
SPADE                                         & \textbf{74.55} & \textbf{95.72} & 33.94          & \textbf{55.02}    & 92.93      & \textbf{51.18}   & 42.38          & \textbf{75.48} & 124.96          \\ \hline
SEAN                                          & 72.63          & 95.28          & \textbf{22.41} & 52.52    & 92.53      & 52.62   & 41.41          & 74.93          & 127.38          \\ \hline
\textbf{Ours}                                 & 73.89          & \textbf{95.72} & 25.49          & 53.37    & \textbf{93.01}      & 60.45   & \textbf{43.59} & 75.45          & \textbf{119.82} \\ \hline
\end{tabular}}
\label{sota_table}
\end{table}

\begin{table}[t]
\centering
\small
\caption{Quantitative comparison on reconstruction performance and perceptual similarity. Higher PSNR, lower NRMSE, and lower LPIPS indicate better performance.}
\resizebox{\textwidth}{!}{\begin{tabular}{l||c|c|c||c|c|c||c|c|c}
\hline
\multicolumn{1}{c||}{\multirow{2}{*}{Method}} & \multicolumn{3}{c||}{CelebAMask-HQ}               & \multicolumn{3}{c||}{Cityscapes}                  & \multicolumn{3}{c}{CMP Facades}                 \\ \cline{2-10} 
\multicolumn{1}{c||}{}                        & PSNR           & NRMSE          & LPIPS          & PSNR           & NRMSE          & LPIPS          & PSNR           & NRMSE          & LPIPS          \\ \hline \hline
Pix2PixHD                                     & 15.78          & 0.451          & 0.359          & 16.25          & 0.447          & 0.393          & 12.23          & 0.507          & 0.441          \\ \hline
SPADE                                         & 14.35          & 0.475          & 0.373          & 17.35          & 0.458          & 0.380          & 12.93          & 0.575          & 0.420          \\ \hline
SEAN                                          & \textbf{18.54} & \textbf{0.248} & 0.274          & 20.08          & 0.348          & \textbf{0.331} & 14.36          & 0.459          & 0.402          \\ \hline
\textbf{Ours}                                 & 18.22          & 0.263          & \textbf{0.255} & \textbf{20.83} & \textbf{0.325} & 0.349          & \textbf{15.26} & \textbf{0.408} & \textbf{0.392} \\ \hline
\end{tabular}}
\label{sota_table2}
\vspace{-0.3cm}
\end{table}

\subsection{Comparisons with State-of-the-art Methods}
\begin{figure}
    \centering
    \includegraphics[width=\linewidth]{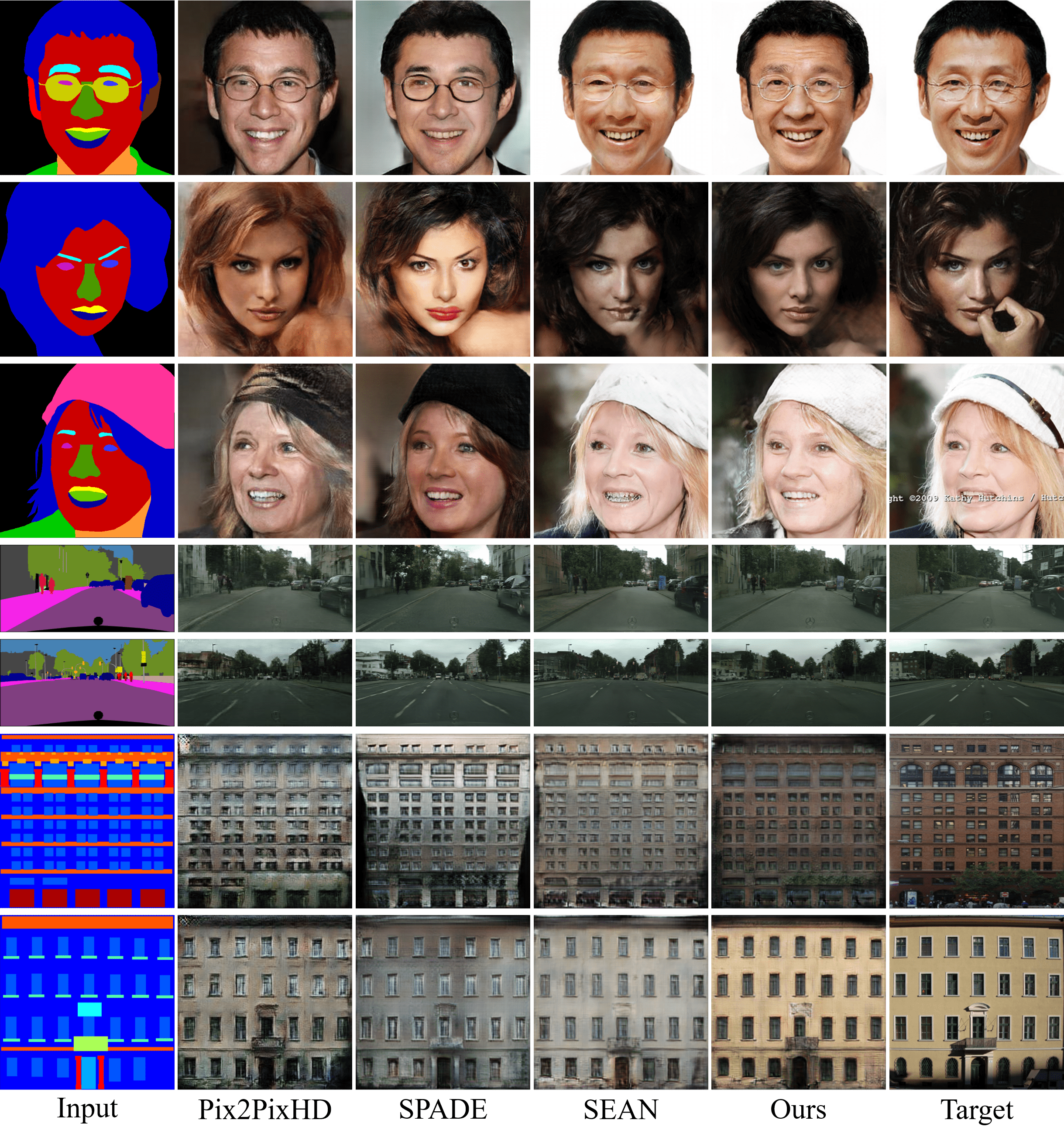}
    \vspace{-0.8cm}
    \caption{Qualitative comparison of semantic image synthesis with state-of-the-art methods on three datasets.}
    \label{SOTA}
\end{figure}

\textbf{Quantitative comparisons.} We quantitatively compare SuperStyleNet with the state-of-the-art ones on semantic segmentation, generation, reconstruction performance, and perceptual similarity. First, we select Pix2PixHD \cite{wang2018high}, SPADE \cite{park2019semantic}, and SEAN \cite{zhu2020sean} as current state-of-the-art methods for comparisons. In Tables \ref{sota_table}, it is obvious that SuperStyleNet generally outperforms these state-of-the-art ones in maintaining segmentation masks. It indicates that all objects including local ones are successfully reconstructed and recognized as these correct classes. However, SuperStyleNet performs slightly worse than SEAN in CelebAMask-HQ in terms of reconstruction metrics, while SuperStyleNet shows better performance in both Cityscapes and CMP Facades as shown in Table \ref{sota_table2}. This is because SuperStyleNet infers personal characteristics (i.e., genders and ages) from segmentation masks due to the lack of the style encoding network. Therefore, SuperStyleNet yields failure cases as shown in Fig. \ref{failure}, which causes degradation of the reconstruction performance. Overall, these results indicate that our SuperStyleNet structure is effective in preserving segmentation masks while generating high-fidelity synthesized images.
\\
\textbf{Qualitative comparisons.} To validate the effectiveness of the proposed method in terms of visual quality, we compare SuperStyleNet with the aforementioned state-of-the-art methods. As shown in Fig. \ref{SOTA}, SuperStyleNet generates high-quality synthesized images on all datasets compared with other methods. Concretely, SuperStyleNet is effective in reconstructing occluded local areas by glasses, hands, or characters on CelebAMask-HQ due to the help of SPSE. Furthermore, small-scale objects, such as human, car, and traffic lights, are well synthesized in Cityscapes. 
Despite injecting style information, SEAN suffers from color distortion when training on the small-scale dataset like CMP facades.
In contrast, SuperStyleNet accurately recovers style information in each semantic layouts matched with its ground-truth. This is because SPSE directly utilizes color values in the source images as style information for style mapping. Overall, SuperStyleNet achieves better preservation of semantic layouts while generating realistic synthesized images compared with the state-of-the-art methods. More comparisons and synthesized images are shown in the supplementary material.

\begin{figure}
    \centering
    \includegraphics[width=\linewidth]{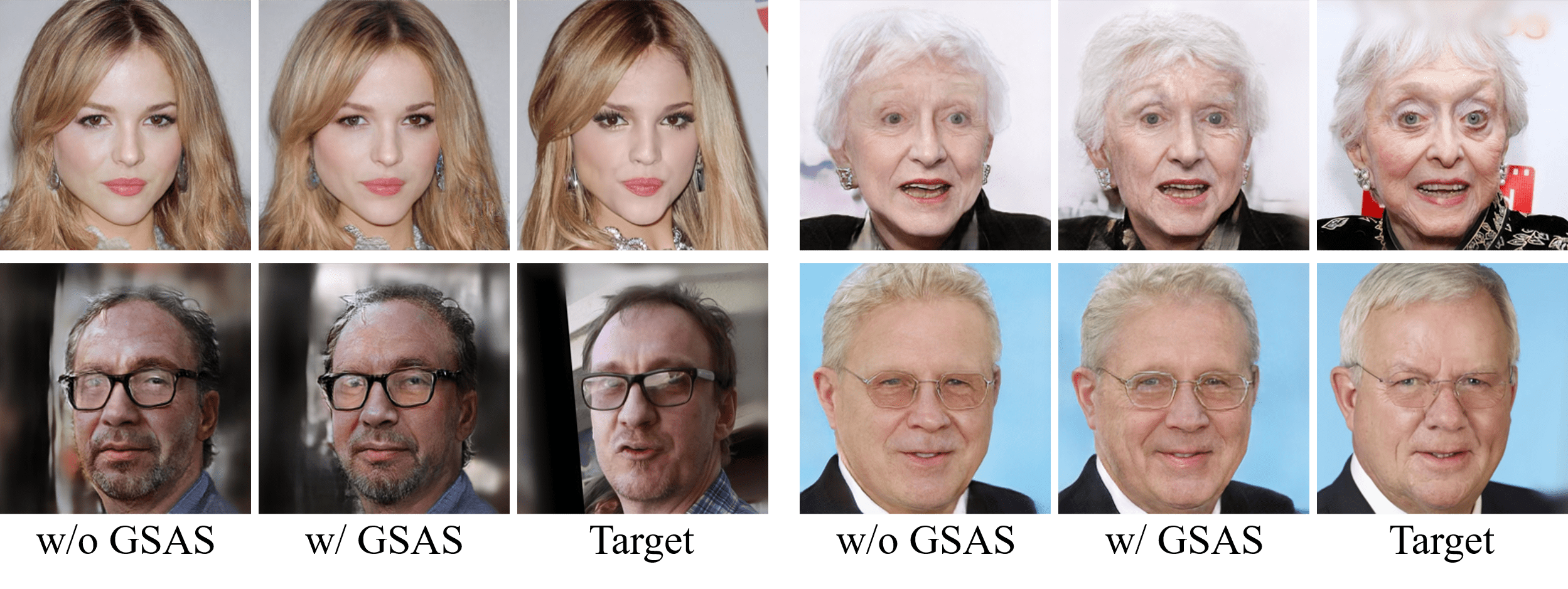}
    \vspace{-0.8cm}
    \caption{Effects of GSAS on the visual quality.}
    \label{ablation}
\end{figure}

\begin{figure}
    \centering
    \includegraphics[width=0.95\linewidth]{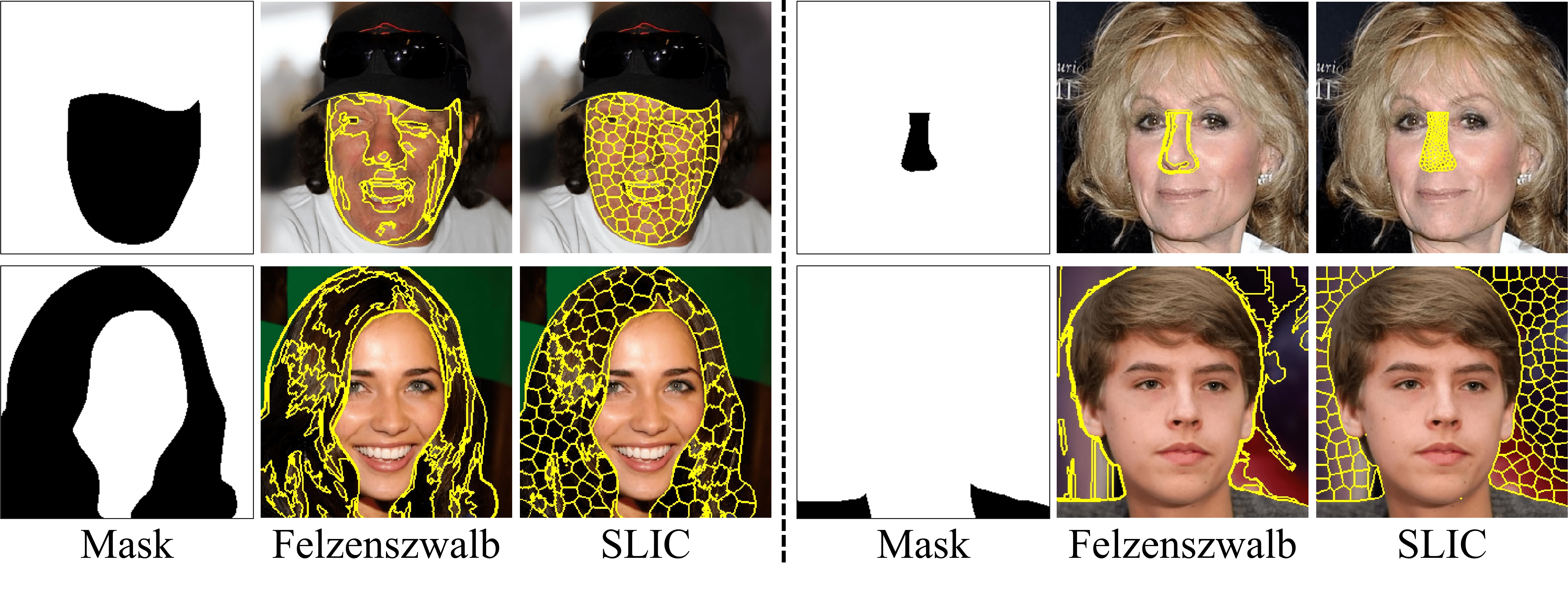}
    \vspace{-0.5cm}
    \caption{Visualization of superpixel segmentation using Felzenszwalb \etal \cite{felzenszwalb2004efficient} and SLIC \cite{achanta2012slic}. Both methods cluster pixels in given semantic layouts into superpixels.}
    \vspace{-0.5cm}
    \label{superpixel_comp}
\end{figure}

\subsection{Ablation Study}
\vspace{-0.3cm}
\textbf{Effectiveness of SuperStyleNet.} For validating the effectiveness of SuperStyleNet, we adopt generator and discriminator of SEAN without a style encoding network as baseline that is regarded as the state-of-the-art model. SPSE substitutes for the style encoding network to extract style information from source images. It can be observed that the segmentation performance is improved from 72.63/95.28 to 73.91/95.58 in mIoU and pixel-wise accuracy on CelebAMask-HQ. Furthermore, its non-parametric operation allows SuperStyleNet to reduce 1.6M parameters while the style encoder increases the inference time around 3s/image due to the CPU processing. In addition, we provide quantitative and qualitative analyses to explore the effects of GSAS. When GSAS is embedded in our network, the reconstruction performance is improved from 18.15/0.285 to 18.22/0.263 in both PSNR and NRMSE on CelebAMask-HQ, while maintaining the segmentation performance. Furthermore, the visual quality is also enhanced as shown in Fig. \ref{ablation}. Specifically, artifacts and details are recovered in the synthesized images due to its inference of spatial representations.
\\
\textbf{Variations of SPSE.} 
SPSE utilizes a superpixel algorithm to encode an input image into style vectors, thus it is mainly affected by the superpixel algorithm and its variations. 

1) \textit{Selection of the superpixel algorithm}: We consider that the number of superpixels should be controllable to obtain the same length of style codes extracted from the same categories in different images. However, conventional methods \cite{felzenszwalb2004efficient, vedaldi2008quick} cannot control the number of superpixels. It incurs information imbalance of style codes in each object since style information is decided by the number of superpixels. Moreover, these methods irregularly encode pixels into superpixels as shown in Fig. \ref{superpixel_comp}. On the other hand, SLIC \cite{achanta2012slic} and LSC \cite{li2015superpixel} not only initialize superpixel centers with uniform distributions but also control the number of superpixels. Thereby, we adopt SLIC as the superpixel segmentation algorithm for SPSE, which shows the faster processing than LSC.

2) \textit{The number of superpixel centers}: This factor determines the amount of information to represent encoded objects. It indicates that the larger number of superpixel centers $k$ is capable of more containing details of objects and diverse color information. Therefore, the generator is able to better reconstruct synthesized images when $k$ is larger. To fully encode the local ones into k-vectors, we set $k$ to the maximum value that does not exceed the number of pixels on the smallest object across datasets. Thus, we empirically set $k$ to 128. To validate it, we conduct experiments on CelebAMask-HQ by changing the parameter $k$.  When $k$ is reduced from 128 to 32, PSNR and mIoU decrease from 18.22 to 17.97 and 73.89 to 69.19. Furthermore, both performances are still lower although we increase $k$ to 64.

\subsection{Style Mixing} 
\begin{figure}
    \centering
    \includegraphics[width=\linewidth]{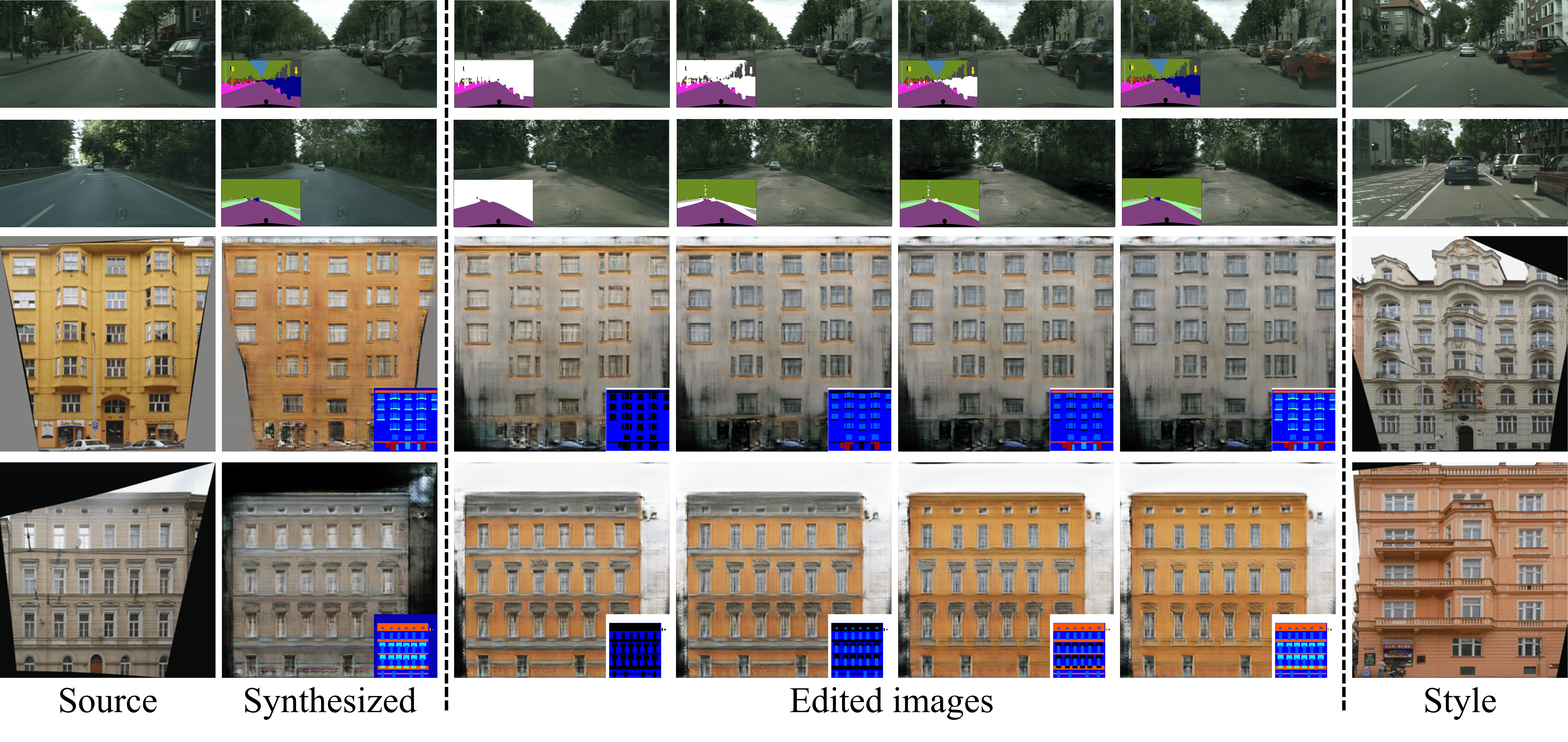}
    \vspace{-0.8cm}
    \caption{Image editing per semantic region. The style vectors are replaced from source to style images on given segmentation masks.}
    \vspace{-0.3cm}
    \label{edit}
\end{figure}

\begin{figure}
    \centering
    \includegraphics[width=\linewidth]{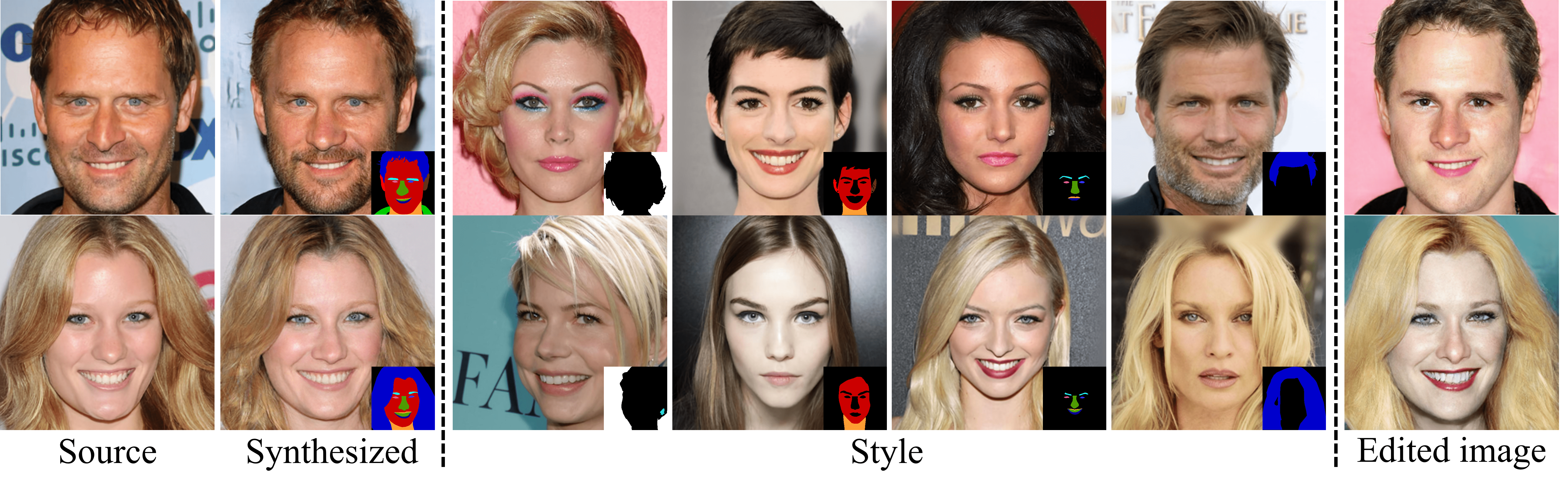}
    \vspace{-0.8cm}
    \caption{Style mixing with multiple style images on given segmentation masks.}
    \vspace{-0.5cm}
    \label{mix}
\end{figure}

\vspace{-0.3cm}
SuperStyleNet is capable of editing images per semantic region or mixing styles of multiple images. To achieve both image manipulations, we change input style vectors from source to style images on given segmentation masks while other vectors retain source ones as shown in \ref{Main} and \ref{edit}. Furthermore, we mix multiple styles so as not to overlap semantic classes as illustrated in \ref{mix}. As can be seen from the figures, the edited images maintain textures and structures of the source images while changing styles by referring to the style images with given segmentation masks. 

\section{Conclusion}
\vspace{-0.3cm}
We propose superpixel based parameter-free style encoding (SPSE) for image synthesis to evenly extract local and global style codes from the original image. To be specific, SPSE clusters pixels in a given image to yield superpixels as style codes per semantic region. Then, the graphical self-attention strategy (GSAS) interprets hidden representations between superpixels to capture spatial relationships. Consequently, SPSE and GSAS facilitate our network to generate high-quality synthesized images matched with target images while preserving semantic layouts. Furthermore, it benefits from reconstruction in occluded regions and small-scale objects. However, the proposed SuperStyleNet yields failure cases in the inference of personal characteristics. In our future work, we will investigate this problem.

\section*{Acknowledgements}
This research was supported by the National Research Foundation of Korea (NRF) grant funded by the Korea government (MSIT) (No. 2020R1A2C1012159).

\bibliography{bmvc_review}

\newpage
\section*{\textit{Supplementary Material}}
Thank you for reading the supplementary material, in which we introduce more details of experiments and SPSE. Moreover, we provide qualitative comparisons and synthesized results.
\vspace{-0.3cm}
\begin{figure}[!h]
    \centering
    \includegraphics[width=\textwidth]{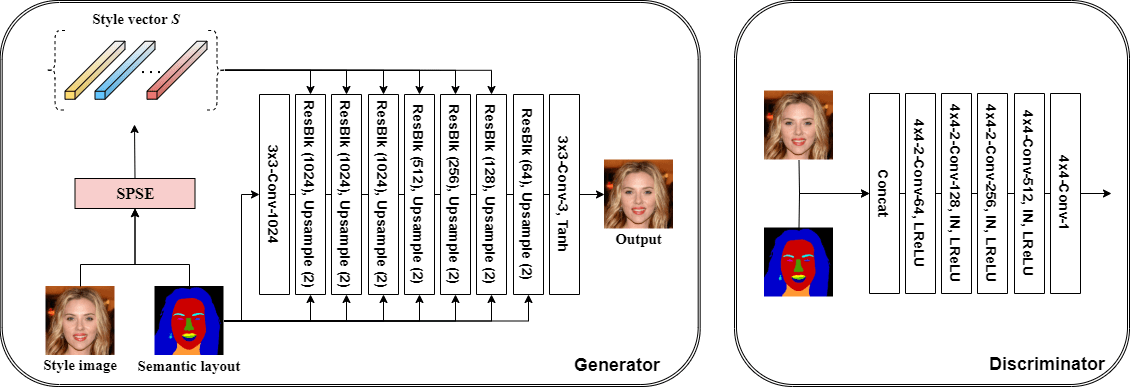}
    \vspace{-0.5cm}
    \caption{Whole framework of SuperStyleNet. (\textit{left}) Structure of the generator. Details of ResBlk are described in Fig. \ref{structure}. (\textit{right}) Structure of the discriminator.}
    \vspace{-0.7cm}
    \label{structure_sup}
\end{figure}
\section*{A. Additional Implementation Details}
\label{A: details}
\textbf{Generator}: Adopting the SEAN generator \cite{zhu2020sean}, the architecture of our generator is comprised of a series of the residual blocks with nearest neighbor upsampling as shown in Fig. \ref{structure_sup}. We use $8\times$ downsampled semantic layouts as input of the generator. The style vectors are generated by SPSE, and they are fed into each residual block excluding the last one. In contrast, the semantic layouts and noise are fed into all residual blocks.
\\
\\
\textbf{Discriminator}: Following the previous works \cite{wang2018high, park2019semantic, zhu2020sean}, we employ two multi-scale discriminators, which consist of convolution layers with spectral normalization \cite{miyato2018spectral}, instance normalization (IN) \cite{ulyanov2016instance}, and Leaky ReLU (LReLU) as illustrated in Fig. \ref{structure_sup}.
\\
\\
\textbf{Learning objectives}: To train the proposed network, the learning objectives are utilized as follows: \\
(1) \textit{Perceptual loss}: We employ the VGG network \cite{simonyan2014very} to calculate the perceptual loss \cite{johnson2016perceptual}. Given a style image $R$ and its corresponding masks $M$, let $\phi_{i}$ be feature maps of the $i$-th layer of the VGG network and $N$ be the number of its layers. Then, the shape of the feature maps is $C_{i}\times H_{i}\times W_{i}$, and the perceptual loss is described as:
\begin{equation}
    L_{percept} = \frac{1}{N}\sum_{i=1}^{N}\frac{1}{C_{i}H_{i}W_{i}}\sum_{c=1}^{C_{i}}\sum_{h=1}^{H_{i}}\sum_{w=1}^{W_{i}}\left \| \phi_{i}(R)_{c,h,w} - \phi_{i}(\textbf{G}(SC, M))_{c,h,w} \right \| _{1},
\end{equation}
where \textbf{G} is the generator, and $SC$ is extracted style codes from the style image.
\\
(2) \textit{Feature matching loss}: Inspired by Pix2PixHD \cite{wang2018high}, we apply the feature matching loss to our network. Let $D_{i}^{k}$ be feature maps of the $i$-th layer of multi-scale discriminators $D^{1}$ and $D^{2}$, and $M_{k}$ be the number of layers in these discriminator, then the feature matching loss is formulated as: 
\begin{equation}
    L_{FM} = \frac{1}{M_{k}}\sum_{i=1}^{M_{k}}\frac{1}{C_{i}H_{i}W_{i}}\sum_{c=1}^{C_{i}}\sum_{h=1}^{H_{i}}\sum_{w=1}^{W_{i}}\left \| D_{i}^{k}(R,M)_{c,h,w} - D_{i}^{k}(\textbf{G}(SC, M),M)_{c,h,w} \right \| _{1}.
\end{equation}
\\
(3) \textit{Adversarial loss}: We adopt the hinge loss term \cite{miyato2018spectral, zhang2019self} for adversarial learning. Then, the learning objective is formulated as:
\begin{equation}
\begin{aligned}
    &L_{GAN}=-\mathbb{E}[\min(0,-1+D_{k}(R,M))]-\mathbb{E}[\min(0,-1-D_{k}(\textbf{G}(SC,M),M))],\\
    &L_{ad}=-\mathbb{E}[D_{k}(\textbf{G}(SC,M),M)],
\end{aligned}
\end{equation}
where $L_{GAN}$ and $L_{ad}$ are for the discriminator and the generator, respectively.

Overall, the final loss function for SuperStyleNet is described as:
\begin{equation}
    L_{total}=\alpha L_{percept} + \sum_{k=1,2} (\beta L_{FM} +  L_{ad}),
\end{equation}
where $\alpha$ and $\beta$ are the weight parameters of individual loss terms, and both are set to 10, respectively.
\vspace{-0.3cm}
\section*{B. Additional Details of SPSE}
\label{B: SPSE}
\vspace{-0.3cm}
\textbf{Algorithm}: In this paper, we propose superpixel based parameter-free style encoding (SPSE) to encapsulate the input image in the style codes as described in Algorithm \ref{algo_spse}. To be specific, we convert the RGB input image $I\in \mathbb{R}^{n\times3}$ to a five-dimensional color and position space ($LabXY$) $\hat{I} \in \mathbb{R}^{n\times5}$, where $n$ is the number of pixels.  After that, we initialize $k$ superpixel centers $S^{0}_{l}\in \mathbb{R}^{k\times 5}$ in a given semantic mask $M_{l}$ using an uniform distribution \cite{irving2016maskslic}, and extract pixels $p_{l}$ from a converted input $\hat{I}$ in the given semantic mask $M_{l}$. Then, we compute the association map $Q\in \{0,...,k-1\}^{n\times1}$ between the pixels and nearest superpixel centers by computing the distance during iteration $t$:
\begin{equation}
    Q_{p_{l}}^{t}=\mathop{\arg\min} _{i\in \{0,...,k-1\}}D(\hat{I}_{p_{l}},S^{t-1}_{i}),
\end{equation}
where $D$ is the Euclidean distance. To update new superpixel centers in each iteration, we take average pixel features of the converted input $\hat{I}$ inside each superpixel cluster $i$:
\begin{equation}
    S^{t}_{i_{l}}=\frac{1}{Z^{t}_{i_{l}}}\sum_{p_{l}|Q^{t}_{p_{l}}=i_{l}}\hat{I}_{p_{l}},
    \label{center}
\end{equation}
where $Z^{t}_{i_{l}}$ is the number of pixels in the superpixel cluster $i_{l}$. After proceeding all iterations, we obtain the final association maps $Q^{\prime}$ and superpixel centers $S^{\prime}$ in the given semantic label. To convert the superpixel centers to the style codes  $SC\in \mathbb{R}^{k\times 3}$, we repeat Eq. \ref{center}  with the original input $I$:
\begin{equation}
	SC_{i_{l}} = \frac{1}{Z_{i_{l}}}\sum_{p_{l}|Q^{\prime}_{p_{l}}=i_{l}}I_{p_{l}}.
\end{equation}
Finally, the style codes are reshaped to $3k$, and interpolated with the desired length $N$ of the style ones.

\begin{algorithm}
    \caption{Superpixel based Parameter-free Style Encoding (SPSE)}\label{algo_spse}
    \hspace*{\algorithmicindent} \textbf{Input}: Image $I$, Semantic labels $L$. \\
    \hspace*{\algorithmicindent} \textbf{Output} : Style codes $SC$ per semantic mask.
    \begin{algorithmic}[1]
    \State{Convert input $I$ to the $LabXY$ space $\hat{I}$.}
    \For{\textit{each semantic label $l\in L$}}
      \State{Initialize superpixel centers $S^{0}_{l}$.}
      \State{Extract pixels $p_{l}$ from $\hat{I}$ in a semantic mask $M_{l}$.}
      \For{\textit{each iteration t}}
      \State{Compute association between each pixel $p_{l}$ and the nearest superpixel $i_{l}$,
      $Q^{t}_{p_{l}}=\mathop{\arg\min} _{i\in \{0,...,k-1\}}D(\hat{I}_{p_{l}},S^{t-1}_{i})$.}
      \State{Compute new superpixel centers, $S^{t}_{i_{l}}=\frac{1}{Z^{t}_{i_{l}}}\sum_{p_{l}|Q^{t}_{p_{l}}=i_{l}}\hat{I}_{p_{l}}$.}
      \EndFor{\textbf{end for}}
      \State{Convert the final superpixel centers to the style code, $SC_{i_{l}} = \frac{1}{Z_{i_{l}}}\sum_{p_{l}|Q^{\prime}_{p_{l}}=i_{l}}I_{p_{l}}$.}
     \EndFor{\textbf{end for}}
  \end{algorithmic}
\end{algorithm}
\vspace{-0.3cm}
\section*{C. Additional Results}
\label{C:results}
\begin{figure}
    \centering
    \includegraphics[width=\textwidth]{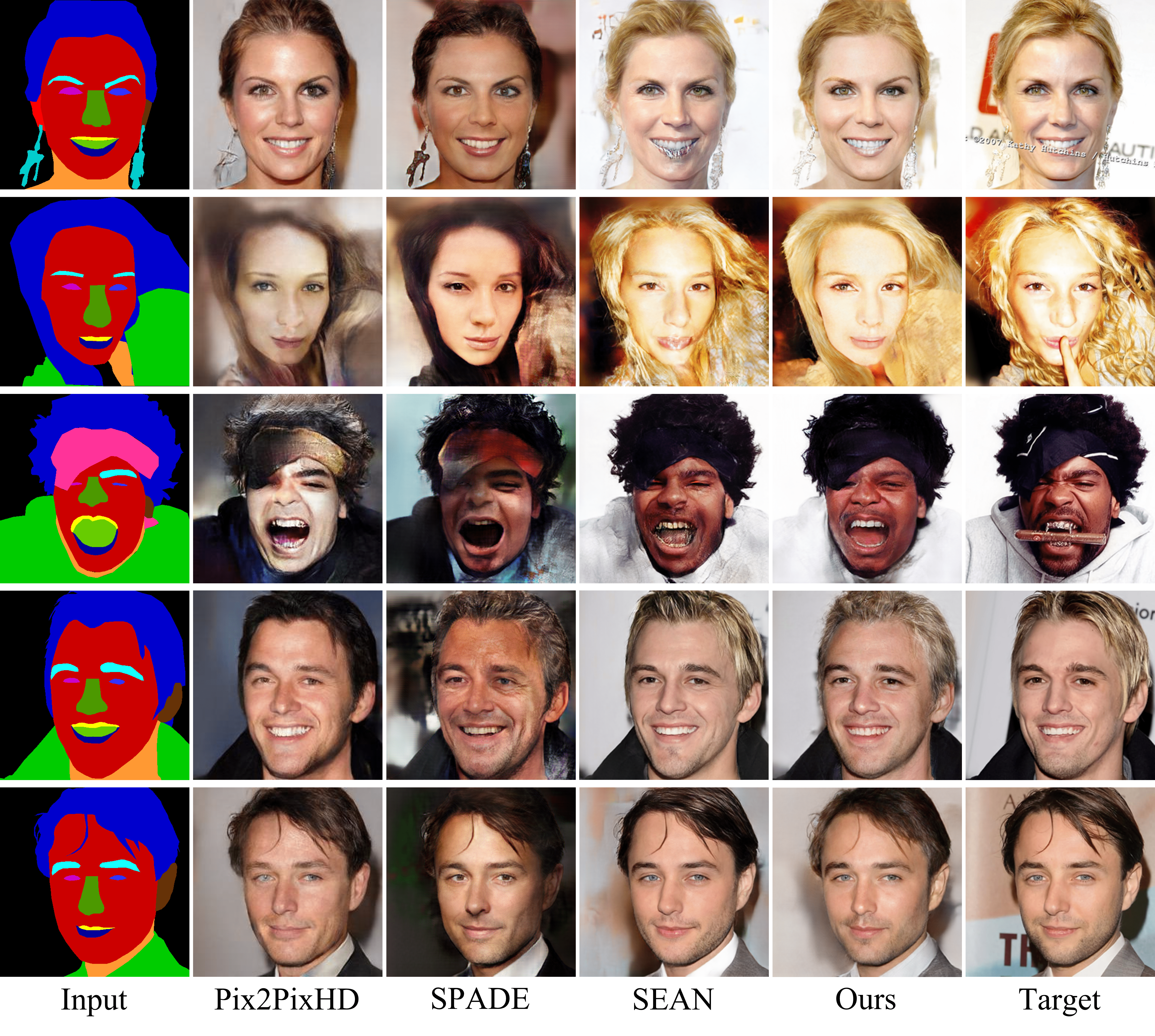}
    \caption{Qualitative comparison of semantic image synthesis on CelebAMask-HQ.}
    \label{celeba_sup}
\end{figure}

\begin{figure}
    \centering
    \includegraphics[width=\textwidth]{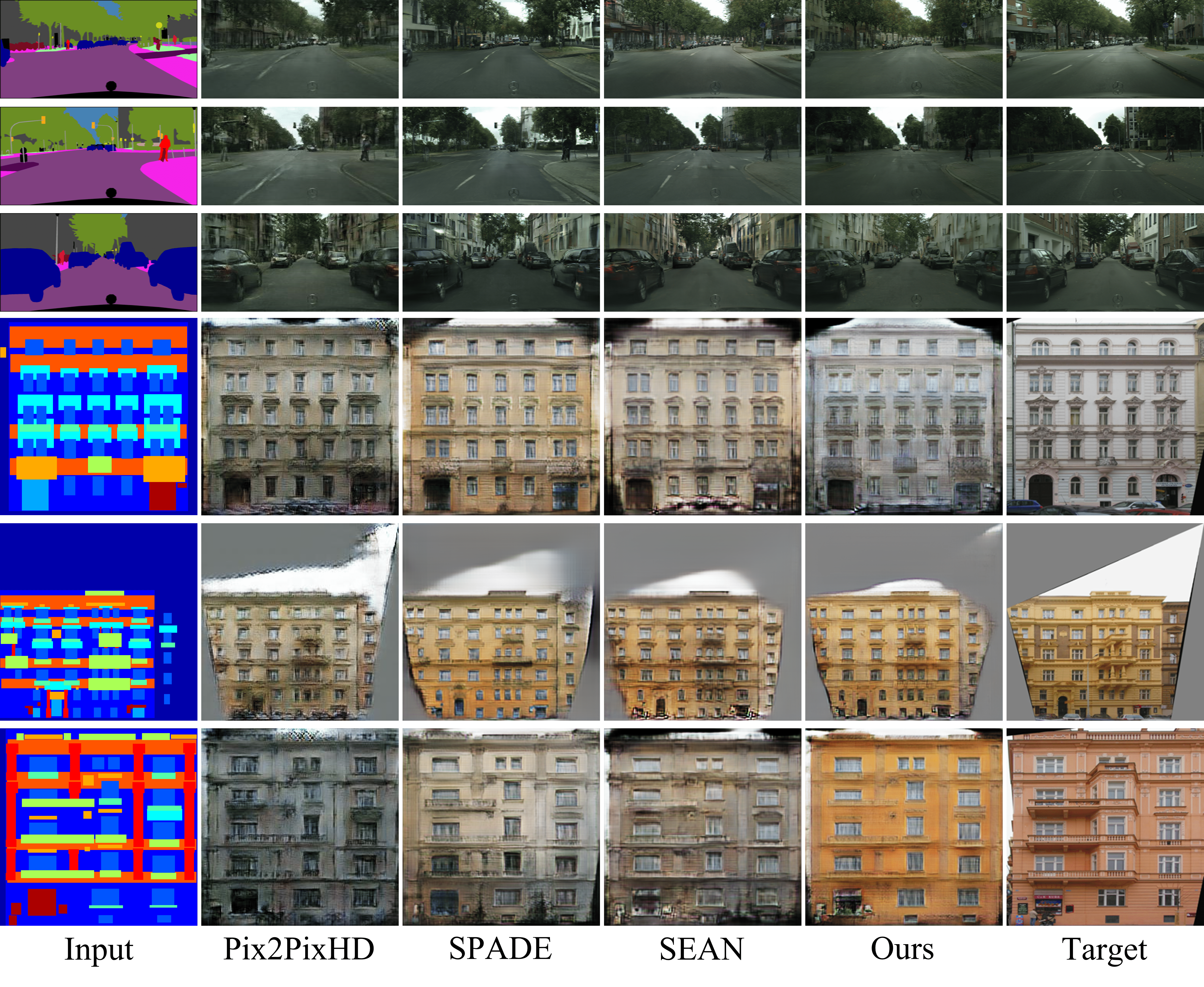}
    \vspace{-0.5cm}
    \caption{Qualitative comparison of semantic image synthesis on Cityscapes and CMP Facades.}
    \vspace{-0.5cm}
    \label{city_faca_sup}
\end{figure}

\begin{figure}
    \centering
    \includegraphics[width=\linewidth]{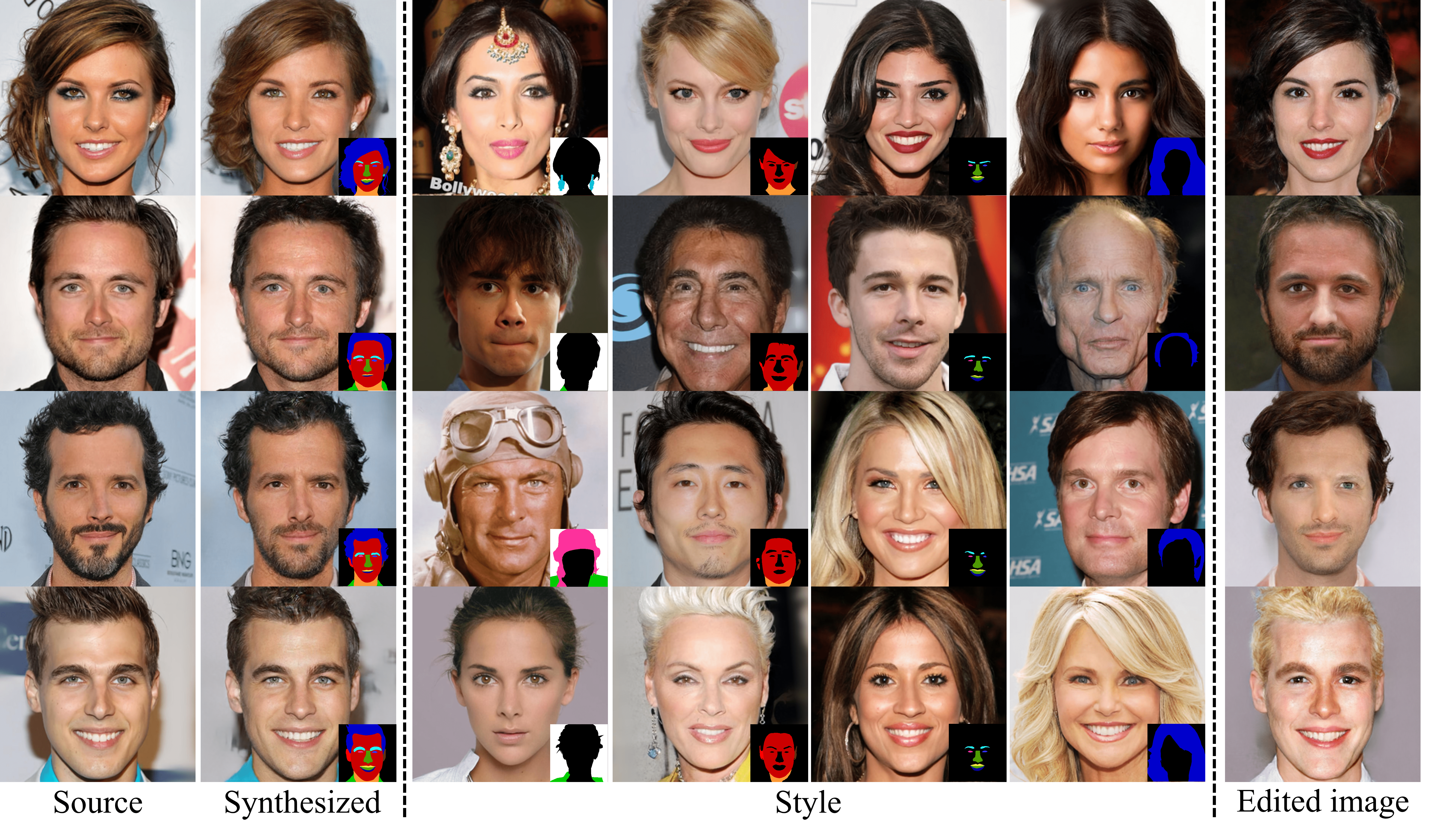}
    \vspace{-0.5cm}
    \caption{Style mixing with multiple style images on given segmentation masks.}
    \vspace{-0.5cm}
    \label{mix_sup}
\end{figure}

\vspace{-0.3cm}
To validate the effectiveness of SuperStyleNet, We provide more qualitative comparison of semantic image synthesis on CelebAMask-HQ, Cityscapes, and CMP Facades in Fig. \ref{celeba_sup} and Fig. \ref{city_faca_sup}. Furthermore, we combine multiple style images on given segmentation masks to generate style mixed images as shown in Fig. \ref{mix_sup}.


\end{document}